\documentclass{article}

     \PassOptionsToPackage{numbers, compress}{natbib}

\usepackage[final]{neurips_ts4h_2022}

\usepackage[utf8]{inputenc} 
\usepackage[T1]{fontenc}    
\usepackage{hyperref}       
\usepackage{url}            
\usepackage{booktabs}       
\usepackage{amsfonts}       
\usepackage{nicefrac}       
\usepackage{microtype}      
\usepackage{xcolor}         
\usepackage{graphicx}
\usepackage{float}
\usepackage{soul}
\usepackage{amsmath,amssymb}
\usepackage{multirow}

\title{PiRL: Participant-Invariant Representation Learning for Healthcare }

\author{%
  Zhaoyang Cao\\
  Department of Computational Applied Mathematics and Operations Research\\
  Rice University\\
  Houston, TX 77030 \\
  \texttt{zc48@rice.edu} \\
   \And
   Han Yu, Huiyuan Yang, Akane Sano \\
   Department of Electrical and Computer Engineering \\
   Rice University \\
   Houston, TX 77030 \\
  \texttt{\{hy29, hy49, akane.sano\}@rice.edu} \\
}

\begin{document}
\date{}

\maketitle

\begin{abstract}

Due to individual heterogeneity, performance gaps are observed between generic (one-size-fits-all) models and person-specific models in data-driven health applications. However, in real-world applications, generic models are usually more favorable due to  new-user-adaptation issues and system complexities, etc. To improve the performance of the generic model, we propose a representation learning framework that learns participant-invariant representations, named PiRL. The proposed framework utilizes maximum mean discrepancy (MMD) loss and domain-adversarial training to encourage the model to learn participant-invariant representations. Further, a triplet loss, which constrains the model for inter-class alignment of the representations, is utilized to optimize the learned representations for downstream health applications. We evaluated our frameworks on two public datasets related to physical and mental health, for detecting sleep apnea and stress, respectively. As preliminary results, we found the proposed approach shows around a 5\% increase in accuracy compared to the baseline.
\end{abstract}

\section{Introduction}

Deep learning models have been developed using time-series data for solving health-related problems. For example, for heart diseases, Oh \textit{et al.} proposed an automated system that combines a convolutional neural network (CNN) and long short-term memory network (LSTM) for the diagnosis of arrhythmia \cite{oh2018automated}. Erdenebayar \textit{et al.} designed a deep neural network, recurrent neural networks, and gated-recurrent unit to distinguish apnea and hypopnea events using an electrocardiogram (ECG) signal \cite{erdenebayar2019deep}. Additionally, for mental health,  Yu and Sano applied semi-supervised learning on leveraging unlabeled data to estimate the wearable-based momentary stress \cite{yu2022semi}. Radhika \textit{et al.} proposed the frameworks that investigate the effectiveness of transfer learning and deep multimodal fusion on CNN stress detection models \cite{radhika2020transfer} \cite{radhika2021deep}.

As shown in the aforementioned studies, the deep learning methods have achieved some promising results in health applications. At the same time, researchers have observed that due to the heterogeneity among data including labels, person-specific models outperform generic models \cite{gjoreski2016continuous,kogler2015psychosocial, nakashima2015stress, nkurikiyeyezu2019effect,schmidt2018introducing, zenonos2016healthyoffice}. For instance, Bsoul \textit{et al.} showed that the accuracy of the subject-dependent sleep apnea classification model is 6\% higher than that of the subject-independent model \cite{bsoul2010apnea}. Moreover, Nath \textit{et al.} \cite{nath2020comparative} showed a performance gap of 22.5\% in accuracy between the subject-dependent and the generic LSTM models in emotion recognition. Although person-specific models have been widely proven to outperform the generic models in health applications \cite{healey2005detecting, koldijk2016detecting, valenza2014revealing}, we cannot neglect its drawbacks. First, person-specific models cannot be easily extended to datasets from new populations the models have not seen yet \cite{nkurikiyeyezu2019effect}. Also, it is expensive to collect enormous datasets from individuals to build person-specific models.

Researchers have explored improving the performance of generic models by introducing person-specific information.
For example, Radhika \textit{et al.} used person specific information in the testing set during the feature extraction and selection \cite{radhika2020transfer, rashid2019times}. Bethge \textit{et al.} utilized MMD loss to impose domain-invariant representations for emotion classification tasks\cite{bethge2022domain}, where each participant had his/her own private encoder with a classifier shared among all. Therefore, even though the performance improvements were observed in the generic models, problems still exist in the studies mentioned above because individual encoders result in high computational costs and difficulties in adapting to other subjects.

In this work, we aim to improve the performances of generic models without introducing person-specific information into the model, thus avoiding the aforementioned drawbacks of the person-specific methods. We propose a representation learning method to learn participant-invariant features, which alleviates the heterogeneous issues by integrating the maximum mean discrepancy (MMD) loss in representation learning to minimize the distribution shifts among features from different subjects. Additionally, we use domain classification loss from domain-adversarial training of neural networks (DANN) architecture, which aimed to blur the participant-distinguishable information among the learned representations, to make label predictor more robust to the target participant \cite{bethge2022domain}. Further, the triplet loss, which aims to learn the label-distinguishable embedding, is used in part of the framework as another constraint to avoid trivial solutions in learned representations. We evaluate the proposed method using two public datasets, including downstream tasks: sleep apnea detection and stress detection. Our results suggest that the proposed method can help improve the model performance significantly compared to the baseline model with only an auto-encoder and a supervised learning model without any constraints.

\section{Methodology}
We propose a representation learning framework that aims to extract participant-invariant representations, named PiRL. The main PiRL framework is visualized in Figure \ref{fig:PiRL} and detailed architecture is shown in Appendix \ref{model_description}. We employ a 1D CNN-based auto-encoder structure as a deep feature extractor from input wearable data. On top of the auto-encoder, an MMD loss and domain classification loss is utilized to constrain the representations from distribution shifts to encourage the learning embedding to be participant-invariant. During training for downstream tasks, we optimize the model with a triplet loss for label-distinguishable representations. The following subsections will introduce the aforementioned components in detail. 

\begin{figure}[!htbp]
    \centering\includegraphics[width=0.96\textwidth,height=5.2cm]{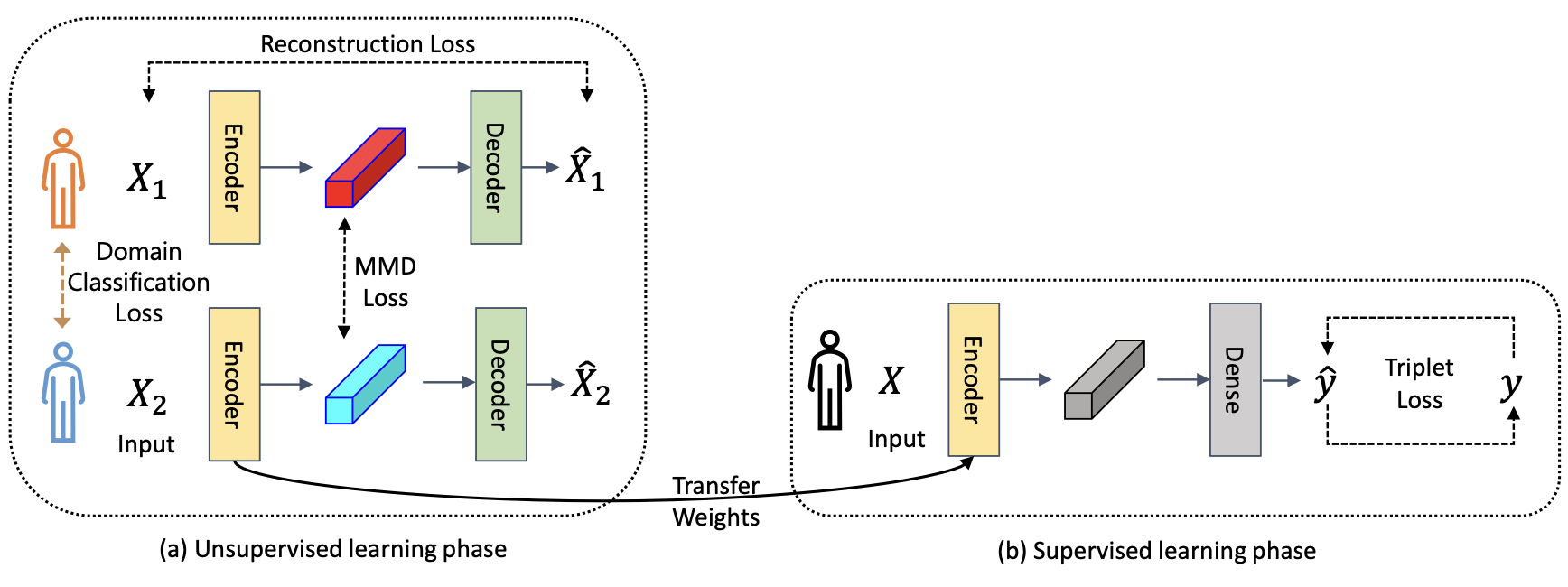}
    \caption{PiRL Network Architecture}
    \label{fig:PiRL}
\end{figure}

\subsection{Representation learning}
We utilize a 1D CNN-based auto-encoder to extract learning representations from raw time-series sequences $X$. The encoder extracts latent representations from input sequences as $e$ with a series of 1D CNN layers; whereas the decoder aims to output the reconstructed signal as $\hat{X}$ from $e$ using up-sampling layers. The objective of the auto-encoder is:
\begin{equation}
    \mathcal{L}_{ae} = \|X - \hat{X}\|_2^2
\end{equation}

\subsubsection{MMD loss} 
To encourage the model to learn the participant-invariant representations, we constrain the model with an MMD loss function, which is widely used in eliminating distribution shifts among different groups of data \cite{gretton2012kernel}.

Given training samples from two different subjects as $X_i$ and $X_j$ with the total number of subjects $N$, the MMD loss can be considered as follows:
\begin{center}
$\mathcal{L}_{mmd}(p,q,\mathcal{H}) = \sum_{i=1}^{N}\sum_{j=1}^{N}\sup\limits_{f \in \mathcal{H},||f||_{\mathcal{H}}\leq1}(
\mathop{\mathbb{E}}\limits_{p(X_i)}[f(X_i)] -\mathop{\mathbb{E}}\limits_{q(X_j)}[f(X_j)])$
\end{center}
where $\textit{p}$ and $\textit{q}$ are the distributions of variable $X_i$ and $X_j$, and $\mathcal{H}$ is the Hilbert space. During the training process, the model is optimized to minimize the distribution distances between the pairs of subjects. Thus, the overall objective of unsupervised learning is:
\begin{equation}
    \mathcal{L} = \mathcal{L}_{ae} + \lambda \cdot \mathcal{L}_{mmd}
\end{equation}
where $\lambda$ is the coefficient of MMD loss and set to 0.2.

 \subsubsection{Domain classification loss} 

A domain classifier from DANN architecture is a potent tool to make the label predictor more robust in target data by 
accomplishing the following adversarial tasks: minimizing the loss of label prediction and maximizing the loss of domain classification. Subsequently, with the addition of a gradient reversal layer before the domain classifier, the overall objective function is the sum of two minimizing problems \cite{ganin2016domain}. In our case, we use an auto-encoder as a feature extractor, label predictor for supervised learning, and implemented a domain classifier in the training process together with reconstruction loss. In order to generate the domain classification, we treat each participant as an independent domain, with the number of domains equal to the number of participants. The corresponding domain classification loss is given below:
\begin{equation}
    \mathcal{L}_{domain} = \|d - \hat{d}\|_2^2
\end{equation}
where $d$ is the one-hot encoding of the source domain and $\hat{d}$ is the one-hot encoding computation of the target domain prediction output. As a result, the overall objective of unsupervised learning is:
\begin{equation}
    \mathcal{L} = \mathcal{L}_{ae} + \lambda \cdot \mathcal{L}_{domain}
\end{equation}
where $\lambda$ is the coefficient of domain classification loss and set to -1.

\subsection{Fine-tuning with triplet loss}
The supervised learning models for downstream tasks are built on top of the learned representations. We design a fully connected dense layer as a classifier that follows the aforementioned CNN-based encoder that is pre-trained with the auto-encoder. Then the supervised learning structure is fine-tuned according to different downstream tasks. Furthermore, we also apply the triplet loss to optimize the representation of the training labels and avoid the trivial solutions learned in the pre-training procedure \cite{schroff2015facenet}. The triplet loss is given below:
\begin{equation}
\label{eq:triplet}
\mathcal{L}_{triplet} = \max (d(a,p)-d(a,n)+margin, 0)
\end{equation}
where $\textit{a}$ represents anchor sample data, 
$\textit{p}$ represents `positive' sample data with the same label from the anchor, $\textit{n}$ represents `negative' sample data with the opposite label against the anchor and the margin is a positive scalar. From the equation \ref{eq:triplet}, we can see that the objective of the triplet loss is to decrease the distance between samples with the same labels and separate the distance between samples with different labels. During the supervised learning phase, the coefficient of triplet loss is set to 0.2.

\section{Results and Discussion}
We tested the proposed PiRL frameworks on two datasets, including CLAS and Apnea-ECG datasets for applications in mental health and physical health.
Appendix \ref{experimental_setting} included detailed information on two datasets and experimental settings such as hyper-parameters. For each supervised prediction model, we pre-trained and fine-tuned the encoder at the beginning of each epoch. We compared the prediction accuracies of the PiRL models against the ones of the baseline (only an auto-encoder and a supervised learning model without any additional constraints) and person-specific models.

\subsection{Stress Detection using Electrodermal
Activity (EDA) with CLAS Dataset}
Table \ref{table: clas} shows the prediction accuracy in all types of supervised learning models. 
The domain classification loss-based model did not show a significant increase of accuracy. The MMD loss-based model showed a higher accuracy of 66.5\% compared to the baseline results (64.3 \%). The prediction accuracy of the triplet loss only and the MMD + triplet loss models both exceeded 70\%. The results illustrated that the model with triplet loss works better in stress prediction than the one with MMD loss-based models. To examine the statistical significance of the accuracy, we conducted an ANOVA (post-hoc: Tukey) test, and the corresponding results are also shown in Table \ref{table: clas}. The prediction accuracy of the baseline framework was treated as the reference group.
We found that both MMD and triplet loss based models have statistically higher accuracy than the baseline framework, but the triplet loss improved the model performance most obviously in stress detection.  
As expected, the person-specific models showed the highest accuracy (86.8\%) but also highest standard deviations(0.189).

\begin{table}[!htbp]
\centering
\caption{Performance in stress detection on CLAS dataset. P-values are calculated by ANOVA (post-hoc: Tukey)}
\label{table: clas}
\begin{tabular}{l|cccccc}
\toprule
      & Baseline & DANN & MMD & Triplet & MMD+Triplet & Person-Specific\\
    \hline
    Accuracy & 64.3\% & 64.5\%& 66.5\% & 70.1\% & 70.6\% &86.8\%\\
     \hline
     SD&0.012 &0.010& 0.014 & 0.011 & 0.010 & 0.189\\
     \hline
      P-value < 0.01& - 
      &$\times$
      &$\checkmark $ &$\checkmark $ & $\checkmark $ &
      $\checkmark $\\
     \bottomrule
\end{tabular}
\end{table}

\subsection{Sleep Apnea Detection using ECG with Apnea-ECG Dataset}
Table \ref{table: ecg} shows the prediction accuracy of apnea detection using supervised learning models. The baseline model obtained a prediction accuracy of 75.2\% with a standard deviation of 0.014. The accuracy of the domain classification loss-based model showed a slight numerical increase but not statistically significant difference. The accuracy of the remaining three PiRL frameworks reached over 79\% and they were all statistically significantly higher than the baseline results with less standard deviation. The best framework for detecting apnea was the combination of MMD and triplet loss since it achieved the highest prediction accuracy. 
 The prediction accuracy of the person-specific model 
was highest which exceed 95\% and statistically higher than the baseline.

\begin{table}[!htbp]
\centering
\caption{Performance in sleep apnea detection on Apena-ECG dataset. P-values are calculated by ANOVA (post-hoc: Tukey)}
\label{table: ecg}
\begin{tabular}{l|cccccc}
\toprule
      & Baseline & DANN & MMD & Triplet & MMD+Triplet & Person-Specific\\
    \hline
    Accuracy & 75.2\% & 75.7\%
    & 79.5\% & 79.1\% & 79.9\% & 95.7\%\\
     \hline
     SD& 0.014 & 0.013 & 0.010 & 0.013 & 0.009 & 0.011\\
     \hline
     P-value < 0.01& - 
      &$\times$
      &$\checkmark $ &$\checkmark $ 
      & $\checkmark $
      & $\checkmark $\\
     \bottomrule
\end{tabular}
\end{table}

\section{Latent Space Visualization}
Visualization of the latent space is often helpful in demonstrating how the representations perform in a reduced-dimensional space. Hence, we used t-distributed stochastic neighbor embedding (t-SNE) to realize the latent space visualization in this study as it is commonly used for dimension reduction and visualization of high-dimensional datasets \cite{van2008visualizing}. We plotted the distributions of learned representations between the baseline and the MMD loss-based models in 2-dimensional coordinates to evaluate the effectiveness of the MMD loss function. 

\begin{figure}[!htbp]
    \centering  \includegraphics[width=0.7\textwidth,height=6.3cm]{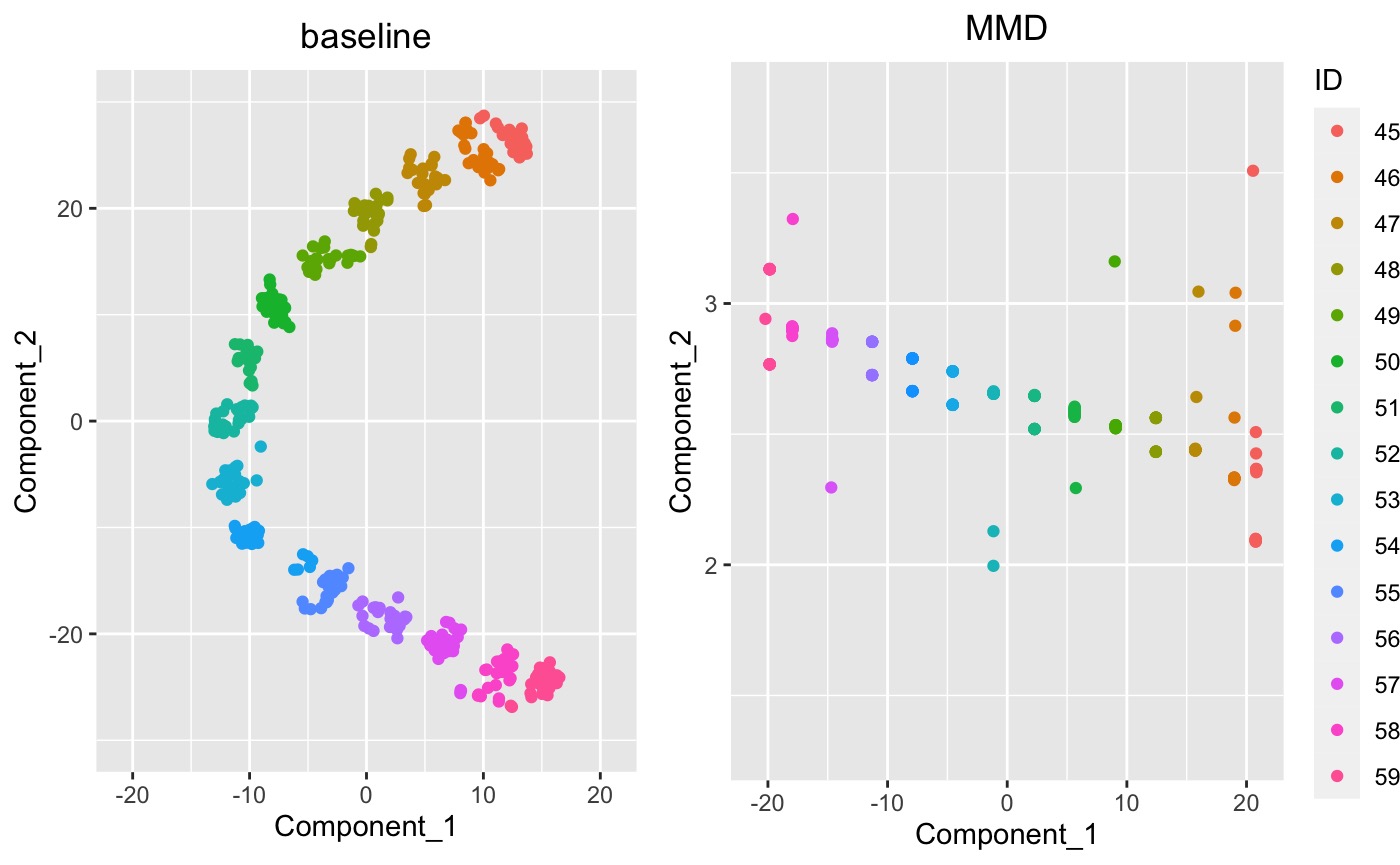}
    \caption{Latent Space Visualization Under MMD }
    \label{fig:latentvismmd}
\end{figure}

Figure \ref{fig:latentvismmd} visualizes the latent space of invariant representations to further explore the influence of MMD loss on CLAS dataset. 
The left baseline figure is the distribution of representations extracted from the auto-encoder without any constrains. Similarly, the right figure is the distribution of representations with MMD loss. 
The range of the second dimensional components on the y axis showed a decrease from [-30,30] (baseline) to [2,4] (MMD) between participants, whereas the first component on the x axis ranged in [-20,20] in both plots. In general, as the representations of each participant spread out in distinct clusters, the latent space visualization of the baseline successfully revealed the heterogeneity across populations. However, for the participant-invariant representation learning, the data from different participants should be entangled and show no collapse in the feature space, but currently, representations on the right plot are still separable during the visualization \cite{akash2021learning}. Therefore, we still need to try some other approaches to better implement the latent space visualization.

\section{Conclusions and Future Work}

In this work, we proposed PiRL, which utilizes MMD and triplet loss for learning participant-invariant representations, to improve the performance of generic health detection models. We evaluated the performance and effectiveness of our framework using two public datasets for mental health and physical health. As preliminary results, we demonstrated that our proposed PiRL outperformed the baseline models and helped generic models achieve better performances. Performance improvement was not observed using DANN technique. The limitation of MMD loss is that it can only shorten the distribution distance in certain dimensions. In future work, we will investigate other approaches to further optimize the representations in data-driven health applications.

\subsection*{Acknowledgments}
This work is supported by NSF \#2047296 and \#1840167.

\bibliographystyle{plain}
\bibliography{reference}

\newpage
\appendix

\section{Model Description}\label{model_description}
\subsection{Unsupervised Learning Phase}
For the CLAS dataset, the encoder
consists of 9 CNN layers, whereas the decoder consists of 8 up-sampling layers.
For the Apnea-ECG dataset, the encoder consists of 12 CNN layers, whereas the decoder consists of 11 up-sampling layers.

\subsection{Supervised Learning Phase}
On both datasets, the encoder structure is same with the structure from unsupervised learning shape. In addition, the dense layer on two datasets both consist of 2 fully connected layers.

\section{Experimental  Setting}\label{experimental_setting}

\subsection{Dataset}
For this study, we investigated two datasets, described as
follows:

\subsubsection{CLAS Dataset}
The CLAS dataset consists of recordings of physiological signals such as ECG, EDA, and Plethysmography (PPG) \cite{markova2019clas}.
EDA time series data were chosen from 62 participants which 58 participants are included in this study. Labels for arousal, valence, and stress were included in the data and assigned based on the stimuli tags for interactive tasks.
We were interested in stress detection using EDA data.

Table \ref{table: datasetpreparationclas} summarizes the EDA data used for the experiments. Training set contains 746 time series sample data with label 0 (`non-stressed') and 247 time series sample data with label 1 (`stressed') from overall 45 participants.
Testing set contains 269 sample data in non-stressed label and 90 sample data in stressed label  form overall 13 participants.

\begin{table}[!htbp]
\centering
\caption{EDA data separation}
\label{table: datasetpreparationclas}
\begin{tabular}{l|ccc}
\toprule
    & \#Participants & \#Non-stressed samples& \#Stressed samples\\
    \hline
    Training Set & 45 & 746 & 247\\
     \hline
     Testing Set& 13 & 269 & 90\\
     \hline
     Total& 58 & 1015  & 337 \\
     \bottomrule
\end{tabular}
\end{table}

\subsubsection{Apnea-ECG Dataset}
The recordings of Apnea-ECG dataset includes continuous ECG signals and sets of annotations for apnea (respiratory signals). The length of the recordings ranges from slightly less than 7 hours to about 10 hours each. The Apnea-ECG dataset consisted of 70 participants was divided into a training set of 35 records (a01 through a20, b01 through b05, and c01 through c10), and a testing set of 35 records \cite{penzel2000apnea}.

Table \ref{table: datasetpreparationecg} summarizes the ECG data separation of the study. The labels of the sample are binary, which 0 represents normal breathing  and 1 represents disordered breathing. 

\begin{table}[!htbp]
\centering
\caption{Apnea-ECG data separation}
\label{table: datasetpreparationecg}
\begin{tabular}{l|ccc}
\toprule
  & \#Participants & \#Normal breathing samples& \#Disordered
breathing samples\\
    \hline
    Training Set & 35 & 10496 & 6514\\
     \hline
     Testing Set& 35 & 10685 & 6548\\
     \hline
     Total& 70 & 21181  & 13062 \\
     \bottomrule
\end{tabular}
\end{table}

\begin{table}[!htbp]
\centering
\caption{Person-specific models data separation}
\label{table: datasetpreparationspecific}
\begin{tabular}{l|ccc}
\toprule
 &  \#Samples in EDA dataset & \#Samples in Apnea-ECG dataset\\
    \hline
    Training Set (Tr1) & 695 & 11907\\
     \hline
     Testing Set (Te1)& 298 & 5103\\
     \hline
     Total& 993 & 17010 \\
     \bottomrule
\end{tabular}
\end{table}

\subsubsection{Data Normalization}
We used min-max normalization on the time series data for the CLAS and Apnea-ECG datasets to ensure that the normalized data had a similar scale. The formula of min-max normalization is given below:
\begin{center}
$x_{scaled}= \frac{x - min(x)}{max(x)-min(x)}$
\end{center}

\subsubsection{Person-specific Models}

The performance of the person-specific models was calculated to compare against the baseline and the proposed PiRL frameworks. We used the original training set of CLAS dataset to obtain the performance of person-specific models. Specifically, for each participant, the original training set was divided into a new training set and testing set with a ratio of 70\%$:$30\%. The person-specific models were trained on the training sets and tested on the test sets to obtain the results. The final prediction accuracy of the person-specific models was recorded as the average accuracy of all participants. 

\subsection{Training Process}
Our tasks are divided into two ways of evaluating representation learning. As for the representation learning, the objection function is mean squared error (MSE) loss between the reconstruction time series data and the original data. We need to obtain robust representations from the auto-encoder that is used in the following step. As for the supervised health condition detection,
CNN-based encoder was pre-trained with the auto-encoder. Then the supervised learning structure is fine-tuned according to different downstream task

In the convolutional auto-encoder, we implemented 100 epochs per trail of training for both the baseline model and regularized model with MMD loss. We shuffled the data, set the batch size and learning rate to 32 and 0.001 at the start of each epoch in CLAS dataset, respectively. The batch size is 256 in Apnea-ECG dataset, which is the only difference between two datasets in training process. We specifically adjusted the weighted coefficient of the MMD loss to be 0.2 in the regularized model to prevent the dominance of the MMD loss. Additionally,
the representation was set to be an 8$\times$1 vector in order to guarantee that the representations we extracted are robust and contain sufficient personalized information. 

In the supervised evaluation model, the parameters such as batch size and learning rate were maintained the same. These four supervised model types—baseline model, MMD loss only, triplet loss only, MMD and triplet loss were being taken into consideration. Noteworthy, the coefficient of the triplet loss was set to be 0.2 as well. Then we pre-trained the auto-encoder, initialized and saved all the representations and parameters in the encoder part. After that, we loaded the saved parameters to the supervised models and fine-tuned them. To get the predicted binary results of labels, we feeded the representations obtained from the encoder to the classifier, which consists of two fully connected layers. The final prediction accuracy was recorded as the mean of 10 trials of training.

\end{document}